\documentclass{article}

 \usepackage[position,preprint]{neurips_2025}

\usepackage[utf8]{inputenc} 
\usepackage[T1]{fontenc}    
\usepackage{hyperref}       
\usepackage{url}            
\usepackage{booktabs}       
\usepackage{amsfonts}       
\usepackage{nicefrac}       
\usepackage{microtype}      
\usepackage{xcolor}         
\usepackage{graphicx} 
\usepackage{enumitem}
\usepackage{amssymb}

\title{When Retrieval Succeeds and Fails: Rethinking Retrieval-Augmented Generation for LLMs}

\author{
  Yongjie Wang $\spadesuit$, Yue Yu $\spadesuit$, Kaisong Song $\clubsuit$, Jun Lin $\clubsuit$, Zhiqi Shen $\diamondsuit$ \\ 
  $\spadesuit$ Alibaba-NTU Global e-Sustainability CorpLab, Nanyang Technological University, Singapore\\
  $\clubsuit$ Tongyi Lab, Alibaba Group, Hang zhou, China \\
  $\diamondsuit$ College of Computing \& Data Science, Nanyang Technological University, Singapore \\
  \texttt{yongjie.wang@ntu.edu.sg, yuyu0022@e.ntu.edu.sg} \\
  \texttt{\{kaisong.sks,linjun.lj\}@alibaba-inc.com, zqshen@ntu.edu.sg} \\
}

\begin{document}

\maketitle

\begin{abstract}
Large Language Models (LLMs) have enabled a wide range of applications through their powerful capabilities in language understanding and generation. However, as LLMs are trained on static corpora, they face difficulties in addressing rapidly evolving information or domain-specific queries. Retrieval-Augmented Generation (RAG) was developed to overcome this limitation by integrating LLMs with external retrieval mechanisms, allowing them to access up-to-date and contextually relevant knowledge. However, as LLMs themselves continue to advance in scale and capability, the relative advantages of traditional RAG frameworks have become less pronounced and necessary. Here, we present a comprehensive review of RAG, beginning with its overarching objectives and core components. We then analyze the key challenges within RAG, highlighting critical weakness that may limit its effectiveness. Finally, we showcase applications where LLMs alone perform inadequately, but where RAG, when combined with LLMs, can substantially enhance their effectiveness. We hope this work will encourage researchers to reconsider the role of RAG and inspire the development of next-generation RAG systems. 
\end{abstract}

\section{Introduction}

Large language models (LLMs) demonstrate extraordinary performance across a wide range of applications, including medical diagnosis \cite{wu2025automated}, behavioral agency \cite{park2023generative,wang2024voyager}, and emotional assistance \cite{wang2025rolerag}. However, relying solely on their static internal knowledge often leads to inaccurate or fabricated outputs in domain-specific or knowledge-intensive tasks, a phenomenon commonly referred to as hallucination \cite{10.1145/3571730,maynez-etal-2020-faithfulness}. To mitigate these limitations, Retrieval-Augmented Generation (RAG) \cite{lewis2020retrieval,mallen2023not} has been proposed to dynamically integrate external, query-relevant knowledge into the generation process, and thus complements the knowledge implicitly encoded in the parameters of LLMs. Therefore, RAG has attracted significant attention and demonstrated notable improvements across a variety of real-world applications.


The success of RAG largely stems from the inherent in-context learning (ICL) capabilities \cite{dong2022survey,olsson2022context} of LLMs, which enable them to condition their outputs on externally supplied evidence and dynamically adapt their reasoning to new contextual inputs. Therefore, advanced RAG research has increasingly focused on enhancing the quality of information provided to LLMs, enabling them to address more complex and knowledge-intensive tasks \cite{edge2024local,gutiérrez2024hipporag,gutiérrez2025ragmemorynonparametriccontinual,leung2025knowledge}. The first line of RAG research focuses on transforming the original queries to facilitate more effective subsequent retrieval \cite{gao-etal-2023-precise,ma-etal-2023-query}; Some other studies aim to fine-tune the embedding model to accurately retrieve the most relevant content \cite{li-li-2024-aoe,zhang2025qwen3}; Recently, much of the research has focused on developing more effective knowledge indexing strategies to facilitate improved retrieval. For example, GraphRAG \cite{edge2024local} and HippoRAG \cite{gutiérrez2024hipporag} incorporate knowledge graphs (KGs) to index the external database, thereby supporting cross-document retrieval; KAG \cite{10.1145/3701716.3715240} integrates the semantic reasoning capabilities of knowledge graphs by traversing along the KG; Agentic RAG \cite{singh2025agentic} combines both LLM agents and RAG methods for addressing complex reasoning tasks that require multi-round LLM revoking.

In the era of increasingly powerful LLMs such as DeepSeek-R1 \cite{guo2025deepseek} and Qwen-3 \cite{yang2025qwen3}, the necessity of RAG is perceived as less compelling, which in turn diminishes recognition of its advances. Although prior RAG research has produced remarkable advances, it is necessary to reconsider whether RAG still effectively complements increasingly powerful LLMs and to identify the key challenges it faces in the current era. In this perspective article, we review recent literature on RAG, first identifying the weaknesses that hinder the reliability and performance of current RAG systems. We then highlight the irreplaceable benefits of RAG, which continue to complement and enhance modern LLMs. The limitations of RAG primarily lie in several aspects: (1) insufficient analysis of what LLMs have already learned, which is essential for determining when to trigger retrieval; (2) inadequate intent analysis in complex questions, which affects the identification of query keywords; (3) unresolved knowledge conflicts within external databases; and (4) a limited understanding of how in-context learning operates within the retrieval-augmented framework. However, we also identify several scenarios that still require the integration of RAG, including knowledge-intensive applications, personalized or private information access, and real-time knowledge integration. By reevaluating these weaknesses and strengths, we aim to explore how RAG system design should evolve alongside the continuous advancement of base LLMs. We hope this work will reaffirm the role of RAG and inspire future research toward more robust, next-generation RAG systems. 

The remainder of this paper is organized as follows. Section \ref{sec:rag} introduces the RAG mechanism and its key modules. Section~\ref{sec:challenges} illustrates the major challenges in RAG and discusses potential avenues for future research. Section~\ref{sec:app} outlines the applications in which LLMs can benefit from RAG. Finally, Section \ref{sec:conclusion} presents the conclusion. 

\section{Anatomy of RAG: Architecture and Key Components}
\label{sec:rag}
In the typical RAG scenario, a user poses a question to LLMs. Owing to intrinsic limitations, LLMs often lack the capacity to provide comprehensive and reliable responses. RAG bridges this gap by retrieving relevant knowledge from external databases. The retrieved content, combined with the user’s query, forms an enriched prompt that enables the LLM to generate more accurate, informative, and contextually grounded answers.

\subsection{Mission of RAG}
Similar to traditional retrieval systems \cite{page1999pagerank,schutze2008introduction}, implementing an ideal retriever that consistently returns tangible content for a given query remains an elusive goal in RAG. For RAG to succeed, retrieval must achieve both comprehensiveness (high recall) and relevance (high precision). In the following sections, we discuss these two objectives separately.
\begin{itemize}[topsep=1pt,leftmargin=10pt]
\item \textbf{High recall.} Recall that RAG is specifically designed to compensate for the limitations of LLMs when addressing knowledge-intensive queries. Therefore, the ultimate goal of the retriever is to ensure that all documents relevant to answering a query—particularly those containing information beyond the LLM’s parametric knowledge—are successfully returned. 
\item \textbf{High precision.} LLMs function as context-conditioned generators, meaning that each retrieved passage serves as a conditioning factor influencing the generation process. That is to say, if the retriever returns noise or irrelevant text, the LLM may incorporate this misleading evidence, producing hallucinations or spurious justifications \cite{chen2024benchmarking,cuconasu2024power}. In addition, if useful information is surrounded by irrelevant content in the middle, the LLM may perform worse—essentially becoming ``lost in the middle \cite{liu2024lost}.'' Therefore, we should minimize the irrelevant content returned. 
\end{itemize}
Generally speaking, retrieval systems face an inherent trade-off between recall and precision. Fetching more documents improves recall but inevitably increases the risk of introducing noise, thereby reducing precision; whereas enhancing precision by suppressing noise may result in the omission of critical information. Achieving an effective balance between recall and precision thus remains a fundamental challenge in retrieval design. 

\begin{figure}
    \centering
    \includegraphics[width=\linewidth]{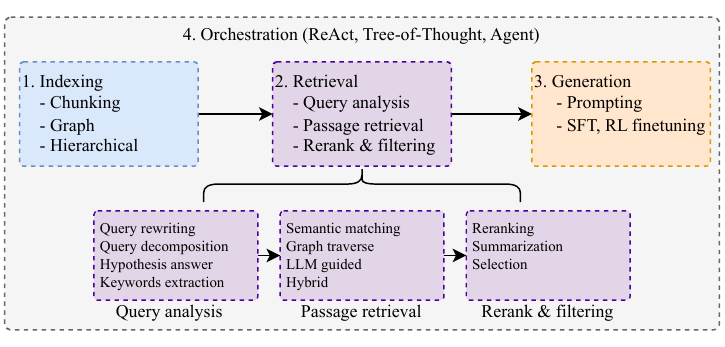}
    \caption{The overall framework of RAG and its four core modules.}
    \label{fig:rag}
\end{figure}

\subsection{Key Modules in RAG}
As illustrated in Figure~\ref{fig:rag}, the RAG system can be divided into four structured modules: indexing, retrieval, generation, and orchestration. Each module represents an essential component of the RAG process, responsible for achieving a distinct high-level objective. In the following, we describe these four modules in detail. 

\textbf{Indexing Module:} At this stage, the indexing module is responsible for organizing and structuring the vast corpus of external sources. A common and straightforward approach is to partition documents into manageable chunks, which are then encoded using representation methods such as BM25 or LLM-based embeddings (e.g., SBERT \cite{reimers2019sentence}, OpenAI text-embedding-3). This simple strategy facilitates efficient retrieval by comparing the similarity between the query embedding and chunk embeddings, enabling seamless integration with LLMs \cite{lewis2020retrieval}.

Despite its effectiveness for semantic similarity--based retrieval, document chunking suffers from two inherent limitations. First, it fails to maintain global coherence across multiple contexts. Second, sentence embeddings are insufficient to capture fine-grained relationships such as causal dependencies or hierarchical taxonomies. To address these challenges, knowledge graph (KG) enhanced RAG has been proposed \cite{edge2024local,gutiérrez2024hipporag,wang2025rolerag}. In this paradigm, external sources are transformed into KGs, where nodes denote entities or concepts and edges encode their relationships. Building on this structure, hierarchical clustering \cite{edge2024local} further organizes entities into multi-level communities, thereby capturing both global coherence and layer-wise conceptual structures.  

One key challenge of the index module is developing an effective indexing system that can accurately match user queries with the most relevant information. Another challenge involves managing heterogeneous data sources—such as PDF, Markdown, HTML, and Word documents.

\textbf{Retrieval Module:} This module bridges the user query and the indexed database, with the goal of retrieving the most relevant data in response to the query. It can be further divided into three sequential steps: \textit{query analysis}, \textit{passage retrieval}, and \textit{reranking and filtering}.

\begin{itemize}[topsep=1pt,leftmargin=10pt]
\item \textbf{Query analysis.} This stage focuses on analyzing users’ questions to infer their intent and generate optimized query terms. The optimized query terms should be both semantically aligned with the user’s original query and sufficiently specific to describe the most relevant information. Key techniques include: Question rewriting \cite{ma2023query}, which involves rephrasing the query from multiple perspectives to improve its expressiveness and coverage; Query decomposition \cite{zhou2023leasttomost}, which aims to break down a complex question into simpler sub-questions following the divide-and-query paradigm; Answer inferring \cite{zhou2024hyqe}, which prompts the LLM to generate hypothetical answers in order to guide retrieval toward semantically relevant content; and keywords extraction \cite{wang2025rolerag}, which identifies salient—often domain-specific—terms from the query and uses them to retrieve specialized or niche knowledge.

\item \textbf{Passage retrieval.} In the RAG framework, retrieval algorithm is designed in accordance with the indexing system. For chunking-based database, retrieval is typically performed by comparing the similarity between the representation of a query (or its expanded form) and the chunks in the indexed database. Common representation methods include sparse encoders (e.g., BM25) and dense embeddings (e.g., SBERT \cite{reimers2019sentence}). More sophisticated multi-stage pipelines hybridize these approaches—for example, by first performing coarse retrieval with BM25 to achieve high recall (e.g., 100+ candidates), followed by semantic retrieval using LLM embeddings to retain the top-K semantically relevant chunks, thus effectively balancing the recall-precision tradeoff. 

Advanced indexing systems further facilitate more sophisticated retrieval. For example, knowledge graph–based indexing enables traversal along the graph structure to identify relevant entities and their interrelationships \cite{edge2024local}. Moreover, community structures formed through graph clustering offer a global and comprehensive understanding that supports question requiring cross-document retrieval. Collectively, these techniques establish a hierarchical retrieval ecosystem that dynamically adapts to query complexity, knowledge domain, and computational constraints.

\item \textbf{Rerank and filtering.} Once relevant passage is retrieved, the rerank and filtering step is designed to refine the results, enhancing precision and filtering out irrelevant noise. Reranking techniques \cite{glass2022re2g} reorder the retrieved content based on the query relevance, retaining only the top similar results. Since feeding all retrieved documents—including irrelevant content—directly into LLMs can dilute attention and cause information distraction, summarization techniques are often employed to retain only the most informative segments, thereby reducing the context length passed to LLMs \cite{gao2025txagent,wang2025rolerag}. 
\end{itemize}

\textbf{Generation Module:} This module instructs LLMs to combines the retrieved data with the user query to produce the final output. Generation performance primarily depends on the LLM’s capability for task understanding and the quality of the retrieved information. Therefore, effective prompt engineering \cite{zhang-etal-2025-prompt-design} is important to ensure that LLM uses the retrieved documents into its intrinsic generation, although there is no universal principle to guide the prompt design. During responses generation, LLMs could encounter conflicting information—either from multiple retrieved sources or between external evidence and their parametric knowledge, and thus LLM should design to suppress inaccurate content from noisy or irrelevant documents \cite{wang2025retrievalaugmented,wang2024resolving}. Correspondingly, some methods \cite{lin2023ra} fine-tune LLMs specifically for the RAG scenario by training them to better distinguish relevant documents from irrelevant ones. 

\textbf{Orchestration Module:} In RAG systems, the above mentioned modules need to be executed sequentially or processed in parallel, depending on system design and task requirements. Therefore, effective workflow planning is essential for improving system efficiency and achieving the desired outcomes. The orchestration module is designed to manage and coordinate interactions among modules and data flows, determining which components to activate based on the specific requirements of a given query. By orchestrating the process, it enables the system to adapt dynamically to varying scenarios, thereby enhancing versatility and efficiency. 

\section{Challenges and Future Directions}
\label{sec:challenges}

Despite recent advances, RAG does not always guarantee superior performance compared with responses generated solely by LLMs \cite{cao2025evaluating,chen2024benchmarking,wu2024how}. In the following, we outline the key challenges in RAG systems and discuss potential solutions to address them.

\subsection{When Should I Retrieve? The Unawareness of LLM Knowledge Boundary}

It is important to note that RAG was originally introduced to compensate for knowledge scarcity in domain-specific tasks. However, a major blind spot of most RAG methods lies in their failure to assess what LLMs already know and what they do not. Instead, these methods often directly apply retrieval over large-scale external sources and report performance improvements without examining necessity or relevance. Meanwhile, LLMs such as DeepSeek \cite{guo2025deepseek} and Qwen3 \cite{yang2025qwen3} have become increasingly powerful, and many fine-tuned models have emerged for specialized domains (e.g., HuatuoGPT-o1 \cite{chen2024huatuogpt} in medical diagnostics and CharacterGLM \cite{zhou-etal-2024-characterglm} in role-playing), leaving limited room for RAG approaches to demonstrate their comparative advantages.

We posit that retrieval is not always necessary for all questions \cite{jeong2024adaptive,jiang2023active} and should instead be triggered adaptively based on the model’s capability. A representative example is provided in prior work \cite{jiang2023active}, where, on the Natural Questions dataset, retrieval-triggering reduced API calls by approximately 40\% without any loss in accuracy. Therefore, it is crucial to first determine whether an LLM can answer a given question solely using its internal knowledge. To this end, uncertainty-based methods can be employed to evaluate prediction variability—for instance, semantic uncertainty \cite{kuhn2023semantic}, self-uncertainty \cite{kang2025scalable}, prediction confidence \cite{leang2025picsar}, and related approaches \cite{guo2025measuring,wang2025response}. RAG is then activated only when the LLM fails to produce a confident prediction on its own.

\subsection{What to Retrieve? The Ineffectiveness of retrieval method}
While RAG excels at fact-oriented questions, it often struggles with complex reasoning tasks (e.g., multi-hop question answering, mathematical reasoning), which require a deep understanding of question intent and the ability to perform step-by-step logical inference. 

Traditional RAG systems typically treat the query as a single unit, extracting lexical features through statistical methods or dense embeddings from LLM-based models. However, simply selecting the Top-K similar chunks cannot adequately capture the extrinsic context or nuanced intent underlying complex queries. For example, a query such as ``If gravity were 10× stronger, how would architectural design evolve?'' might retrieve general physics principles about gravity in reality while overlooking speculative engineering literature. Such retrieved factual data will inevitably lead to an elevated risk of incorrect or hallucinated responses \cite{agrawal2024mindful}. 

Knowledge graph–based indexing facilitates advanced reasoning through traversal along connected edges, it still falls short of meeting the demands of complex reasoning tasks. For instance, frequent entities are often densely connected, which substantially expands the search space and introduces noise during traversal. Current retrieval strategies in KG-RAG generally fall into two categories: (i) K-hop neighborhood, which first identify a seed entity through similarity search and then include its k-hop neighbors, and (ii) LLM-guided search, which leverage language models to evaluate the plausibility of candidate paths. Both approaches exhibit notable limitations—the former often introduces irrelevant entities due to uncontrolled expansion, while the latter is computationally expensive and prone to inconsistency. In addition, for such static graph structures, it is difficult to handle queries involving temporal information, such as identifying the initial, intermediate, and final steps of an execution process. As a result, retrievers struggle to consistently identify the most relevant paths from the KGs for complex reasoning questions.

These retrieval failures can often be attributed to two factors: (1) misinterpretation of the user query, and (2) ineffectiveness of the retrieval method, which together lead to retrieved content that lacks contextual relevance or appropriateness. To facilitate effective retrieval, agent-based frameworks have been introduced to analyze complex reasoning tasks and decompose them into multiple sequential or parallel steps. Within these frameworks, RAG is collaborated with active agents to adaptively retrieve from external knowledge, and thus overcome the challenges of traditional RAG methods, giving rise to the concept of Agentic RAG \cite{singh2025agentic}. Despite the progress in Agentic RAG, future research should continue to focus on achieving a deeper understanding of user intent and developing a unified paradigm capable of adapting to diverse and complex tasks.

\subsection{What Should I Trust? On the Risks of Unverified Data Sources} 
RAG is designed to mitigate factual errors by incorporating external knowledge during the indexing process. However, most RAG methods explicitly assume that external knowledge is inherently reliable and trustworthy, without additional verification \cite{edge2024local,lewis2020retrieval}. In real life, factual inaccuracies often present in retrieval databases. For instance, even the widely used medical database PubMed has been shown to contain fraudulent data and publications, raising concerns about their prevalence and impact \cite{nato2024fraud}. These issues highlight the importance of curating high-quality knowledge bases—such as PrimeKG \cite{chandak2023building}—to fulfill the requirements of LLM-based RAG systems. As tool-augmented agents become increasingly prevalent \cite{gao2025txagent}, constructing high-quality, retrieval-efficient databases tailored for tool integration emerges as a promising avenue for future investigation.

\subsection{How does RAG Work? Linking Retrieval to Mechanism of In-Context Learning}

RAG provides rich, relevant external context that in-context learning (ICL) can leverage to generate higher-quality outputs. However, the mechanisms for resolving conflicts between retrieved evidence and the model’s parametric memory remain unclear, often leading to unpredictable behavior that undermines RAG’s effectiveness. Recent benchmarking studies \cite{huang2025trust} show that when exposed to either correct or incorrect snippets, LLMs tend to rely heavily on retrieved content regardless of its veracity. This underscores the importance of accurate retrieval; yet, even with perfect sources, LLMs may still produce incorrect responses. Such limitations impose an inherent upper bound on RAG’s achievable performance, and the factors determining this bound remain poorly understood. 

To fully exploit the power of RAG, it is essential to understand the mechanisms of ICL \cite{olsson2022context} for anticipating and governing model behavior in long-context settings with explainability techniques \cite{singh2024rethinking,wang2024survey,wang2024gradient}. In particular, we seek to characterize how information flows between the question and retrieved knowledge—for example, which attention heads mediate grounding, how evidence competes with parametric priors, and when positional or recency biases dominate—using tools such as attention rollout/flow, causal tracing and patching, representation probing \cite{wang2025response}, and token-level attribution.

\subsection{RAG vs Long-context LLM}

Long-context LLMs (e.g., GPT-4 \cite{achiam2023gpt}, Claude 3) are capable of processing substantially longer contexts—ranging from hundreds of thousands to over one million tokens—thereby mitigating the reliance on external retrieval. These models are particularly well-suited to tasks such as reasoning over lengthy documents (e.g., books) or multi-document compilations, where ingesting full texts directly is preferable to retrieval.  

However, long-context LLMs also encounter several challenges \cite{bai-etal-2025-longbench,wang-etal-2024-leave}. First, long-context LLMs must suffer from persistent knowledge cutoffs that limit their access to recent information. Second, the quadratic scaling of attention mechanisms leads to substantially higher inference costs. Third, extending the context window increases the likelihood of introducing irrelevant or noisy information. Lastly, as the context length grows, the availability of high-quality training and evaluation data diminishes rapidly, causing long-context LLMs to perform less effectively in practice than reported in controlled benchmarks.

Both approaches address long-document processing but exhibit distinct advantages. Long-context LLMs are more effective when evidence is evenly distributed across multiple documents, whereas RAG performs better in tasks with sparse evidence, assuming accurate retrieval. Additionally, RAG enables access to up-to-date and private information without retraining. A unified framework that integrates RAG with long-context LLMs can leverage their complementary strengths—precise factual retrieval and holistic cross-document reasoning—yielding greater reliability and robustness than either approach alone. 

\section{RAG Applications}
\label{sec:app}

Although recent LLMs demonstrate strong performance across a wide range of tasks, their inherent limitations continue to drive research on RAG, especially in the following fields.  
\begin{itemize}[topsep=2pt,leftmargin=12pt]
\item \textbf{Knowledge-Intensive Applications.} When tasked with knowledge-intensive applications such as drug dosing or rare disease diagnostics, LLMs often perform sub-optimally. In such cases, RAG is particularly powerful, as it enables access to high-quality domain-specific databases and helps mitigate the limitations of parametric memory. Using the retrieved content, the LLM can ground its responses in authoritative evidence, generate more accurate and trustworthy outputs.
\item \textbf{Private Knowledge Management.} Another important application of RAG lies in leveraging private or proprietary knowledge sources, such as enterprise documentation and personal notes. LLMs cannot memorize such data due to the wide diversity and restricted accessibility. In these scenarios, RAG enables customized and secure knowledge retrieval, ensuring that LLMs generate responses aligned with organizational or individual needs while preserving data privacy. A typical example is the multi-turn conversation where the LLM acts as an agent interacting with users.  The conversational history serves as a retrieval database to provide personalized context, thereby enabling more coherent and engaging dialogue. 
\item \textbf{Real-Time Knowledge Integration.} RAG is also effective in domains where knowledge evolves rapidly, such as news, financial markets, and regulatory updates. By continuously retrieving up-to-date information, LLMs can function as information extractors and summarizers, generating responses for questions that reflect the latest developments. 
\end{itemize} 

\section{Conclusion}
\label{sec:conclusion}
As large language models (LLMs) continue to advance in scale and capability, the relative advantages of traditional RAG frameworks have become less pronounced, reshaping RAG’s role in the evolving LLM landscape. This paper provides a systematic review of RAG and its core modules, followed by an analysis of key challenges that limit their effectiveness. Addressing these limitations is essential for ensuring that RAG systems remain robust, adaptive, and complementary to next-generation LLMs. Finally, we outline application domains where RAG remains indispensable for mitigating LLM inefficiencies, underscoring the need for close collaboration between RAG and LLMs. 

\bibliographystyle{plainnat}
\bibliography{reference}
\end{document}